\documentclass{article} 
\usepackage{iclr2021_conference,times}

\usepackage{amsmath,amsfonts,bm}

\def\eqref#1{equation~\ref{#1}}

\def\1{\bm{1}}

\DeclareMathAlphabet{\mathsfit}{\encodingdefault}{\sfdefault}{m}{sl}
\SetMathAlphabet{\mathsfit}{bold}{\encodingdefault}{\sfdefault}{bx}{n}

\usepackage{hyperref}
\usepackage{url}
\usepackage{graphicx}
\usepackage{subcaption}

\def\bphi{{\boldsymbol\phi}}
\def\x{\mathbf x}
\def\y{\mathbf y}

\def\w{\mathbf w}
\def\z{\mathbf z}

\newcommand{\bl}{\color{black}}
\newcommand{\sld}{\color{black}}

\title{Scaling Deep Networks with the Mesh Adaptive Direct Search algorithm}

\author{Dounia Lakhmiri\\
GERAD and Polytechnique Montr\'eal\\
Montr\'eal, Qc, Canada\\
\texttt{dounia.lakhmiri@polymtl.ca}
\And  
Mahdi Zolnouri \\
Huawei Noah's Ark Lab\\
Montr\'eal, Qc, Canada\\
\texttt{mahdi.zolnouri@huawei.com}
\And 
Vahid Partovi~Nia \\
Huawei Noah's Ark Lab\\
Montr\'eal, Qc, Canada\\
\texttt{vahid.partovinia@huawei.com}
\And 
Christophe Tribes\\
GERAD and Polytechnique Montr\'eal\\
Montr\'eal, Qc, Canada\\
\texttt{christophe.tribes@polymtl.ca}
\And 
S\'ebastien Le~Digabel\\
GERAD and Polytechnique Montr\'eal\\
Montr\'eal, Qc, Canada\\
\texttt{sebastien.le.digabel@gerad.ca}
}

\iclrfinalcopy 
\begin{document}

\maketitle

\begin{abstract}
Deep neural networks are getting larger. Their implementation on edge and IoT devices becomes more challenging and moved the community to design lighter versions with similar performance.
Standard automatic design tools such as \emph{reinforcement learning} and \emph{evolutionary computing} fundamentally rely on cheap evaluations of an objective function.
In the neural network design context, this objective is the accuracy after training, which is expensive and time-consuming to evaluate.
We automate the design of a light deep neural network for image classification using the  \emph{Mesh Adaptive Direct Search}
{\sld (MADS) }algorithm,
a mature derivative-free optimization method that effectively accounts for the
expensive blackbox nature of the objective function
to explore the design space, even in the presence of constraints.
{\sld Our tests show competitive compression rates with reduced numbers of trials.}

\end{abstract}

\section{Introduction}


The past decade has seen impressive improvements in the deep neural networks (DNNs) domain that facilitated many technological leaps. DNNs are now applied on increasingly complex datasets and are expected to perform intricate tasks with high precision. However, this comes at the cost of building deeper and more complex networks that require substantial computational resources and, therefore, increase human society's environmental cost. At this stage, searching for new networks has become both essential for technological and scientific advances and expensive enough to prompt more research on how to find small, high-performing networks with a resource-saving mindset.

One possible approach to tackle this issue is to start from a known DNN and reduce its size while losing as few as possible of its original performance.
The popular methods to do so are either by pruning, i.e., keeping the initial architecture and removing some of the network's parameters, or changing the architecture itself through scaling. Low-bit quantization~\citep{courbariaux2015binaryconnect}, or weight sharing~\citep{ullrich2017soft}, or a combination~\citep{han2015deep}, are demonstrated to be effective as well in compressing deep networks. While these methods are not necessarily mutually exclusive, pruning methods are more effective when starting from a low-performing network which stresses the importance of finding a new architecture~\citep{blalock2020state}. Neural architecture search (NAS)~\cite{elsken2018neural} seems to be a necessary and costly step in finding a high-performance neural network, especially for low resource and constraint devices.  

Let $\bphi$ be the vector of hyperparameters that define a network architecture, $\w$ the neural networks weights and $(\x,\y)$ the training data features and labels. We note $f(\bphi, \w\mid\x,\y)$ the performance measure, such as the validation accuracy, of the network after its training; And $g(\bphi)$ the function that reflects the resource consumption, an explicit function such as the number of weights $\w$, or an implicit function such latency on a certain edge device.

Performing an architecture search requires defining a judicious, preferably low-dimensional search space with an efficient method to find a vector $\bphi$ with smallest resource $g$, but with maximum performance $f$ whose evaluation is expensive. Two formulations are proposed to solve this optimization problem. Equation~(\ref{eq:nas_unconstrained}) poses an unconstrained problem with the goal of finding a solution $\bphi \in \Phi$ that maximizes $f$, where $\Phi$ is the search space limited by the box constraints on the components of $\bphi$. The second formulation of~(\ref{eq:nas_constrained}) aims at minimizing the resource function $g$ while maintaining a performance higher or equal to a baseline network defined by the architecture $\bphi_0$ and weights $\w_0$.





\begin{align}
\max_{\bphi \in \Phi}\quad & f(\bphi, \w \mid \x, \y)   \label{eq:nas_unconstrained}\\
\min_{\bphi \in \Phi} \quad        &g(\bphi)~~ \mathrm{subject~to~} ~~ f(\w,\bphi\mid\x,\y ) \geq f(\w_0,\bphi_0\mid \x,\y) \label{eq:nas_constrained}
\end{align}







Several approaches are proposed to solve the maximization of~(\ref{eq:nas_unconstrained}) or the minimization of~(\ref{eq:nas_constrained}), such as grid search (GS) or random search (RS), evolutionary algorithms, reinforcement learning, Bayesian optimization, etc. These methods managed to improve on the state of the art, yet they often ignore essential properties of these optimization problems. Exploring the search space with non-adaptive approaches such as GS or RS, commonly requires many trials and substantial resources, often depleted on inferior architectures.
Similarly, reinforcement learning and evolutionary algorithms need numerous function evaluations before producing quality suggestions. These approaches inherently assume performance evaluation is cheap, which is untrue for training deep networks. 


Here we propose the use of derivative-free optimization (DFO)~\citep{AuHa2017, CoScVibook}, and in particular the Mesh Adaptive Direct Search (MADS) algorithm~\citep{AuDe2006} tailored for expensive blackbox functions. In a DFO setting, the objective function has no derivatives, is costly to evaluate, noisy, stochastic, and its evaluation may fail at some points. Blackbox refers to the cases in which the analytical expression for the evaluating function is unknown. In the NAS context, the blackbox $f$ takes the architecture parameter vector $\bphi$ as the input, trains the network on data $(\x,\y)$, and outputs the validation accuracy.

We present two concrete examples, one where the goal is to compress ResNet-18 on CIFAR-10 by solving the unconstrained problem defined in~(\ref{eq:nas_unconstrained}), and the second aims to compress ResNet-50 On ImageNet by optimizing the constrained formulation of~(\ref{eq:nas_constrained}).

\section{Mesh Adaptive Direct Search (MADS)}\label{sec:nomad}

The Mesh Adaptive Direct Search (MADS) algorithm~\citep{AuHa2017,AuDe2006} is implemented in the {\sf NOMAD} C++ solver~\citep{Le09b} freely
available at {\bl \url{www.gerad.ca/nomad}}.


\begin{figure}
     \setlength{\unitlength}{1.3mm}
       \begin{small}
       \begin{center}
         \begin{minipage}{5.8in}

           \begin{picture}(35,43)(0,0)
             \textcolor{gray}{
               \multiput(4,4)(0,16){3}{\put(0,0){\line(1,0){32}}}
               \multiput(4,4)(16,0){3}{\put(0,0){\line(0,1){32}}}
             }
             \put(20,20){\circle*{.9}}
             \put(22,18){\makebox(0,0){$_{\bphi_k}$}}
             \thicklines
             \multiput(4,4)(0,32){2}{\put(0,0){\line(1,0){32}}}
             \multiput(4,4)(32,0){2}{\put(0,0){\line(0,1){32}}}
             \thinlines
             \textcolor{red}{
               \put(4,20){\circle*{.9}}
               \put(4,20){\line(1,0){16}}
               \put(6,18){\makebox(0,0){$_{t_1}$}}
               \put(20,4){\circle*{.9}}
               \put(20,4){\line(0,1){16}}
               \put(22,2){\makebox(0,0){$_{t_2}$}}
               \put(36,36){\circle*{.9}}
               \put(36,36){\line(-1,-1){16}}
               \put(38,34){\makebox(0,0){$_{t_3}$}}
             }
             \put(4,3){\line(1,0){16}}
             \put(4,2){\line(0,1){1.8}}
             \put(12,0){\makebox(0,0){{\scriptsize $\Delta_k = \delta_k$}}}
             \put(20,2){\line(0,1){1.4}}
           \end{picture}
           \begin{picture}(35,43)(0,0)
             \textcolor{gray}{
               \multiput(4,4)(0,4){9}{\put(0,0){\line(1,0){32}}}
               \multiput(4,4)(4,0){9}{\put(0,0){\line(0,1){32}}}
             }
             \put(20,20){\circle*{.9}}
             \put(22,18){\makebox(0,0){$_{\bphi_k}$}}
             \thicklines
             \multiput(12,12)(0,16){2}{\put(0,0){\line(1,0){16}}}
             \multiput(12,12)(16,0){2}{\put(0,0){\line(0,1){16}}}
             \thinlines
             \textcolor{red}{
               \put(12,16){\circle*{.9}}
               \put(12,16){\line(2,1){8}}
               \put(14,14){\makebox(0,0){$_{t_4}$}}
               \put(20,12){\circle*{.9}}
               \put(20,12){\line(0,1){8}}
               \put(22,10){\makebox(0,0){$_{t_5}$}}
               \put(24,28){\circle*{.9}}
               \put(24,28){\line(-1,-2){4}}
               \put(26,26){\makebox(0,0){$_{t_6}$}}
             }
             \put(4,3){\line(1,0){4}}
             \put(4,2){\line(0,1){1.8}}
             \put(8,2){\line(0,1){1.8}}
             \put(6,0){\makebox(0,0){\scriptsize $\Delta_k$}}
             \put(12,3){\line(1,0){8}}
             \put(12,2){\line(0,1){1.8}}
             \put(20,2){\line(0,1){1.8}}
             \put(16.5,0){\makebox(0,0){\scriptsize $\delta_k$}}
           \end{picture}
           \begin{picture}(35,43)(0,0)
             \textcolor{gray}{
               \multiput(4,4)(0,1){33}{\put(0,0){\line(1,0){32}}}
               \multiput(4,4)(1,0){33}{\put(0,0){\line(0,1){32}}}
             }
             \put(20,20){\circle*{.9}}
             \put(19,19){\makebox(0,0){$_{\bphi_k}$}}
             \thicklines
             \multiput(16,16)(0,8){2}{\put(0,0){\line(1,0){8}}}
             \multiput(16,16)(8,0){2}{\put(0,0){\line(0,1){8}}}

             \thinlines
             \textcolor{red}{
               \put(16,20){\circle*{.9}}
               \put(16,20){\line(1,0){4}}
               \put(14.5,19){\makebox(0,0){$_{t_7}$}}
               \put(24,22){\circle*{.9}}
               \put(24,22){\line(-2,-1){4}}
               \put(26,23){\makebox(0,0){$_{t_9}$}}
               \put(23,16){\circle*{.9}}
               \put(23,16){\line(-3,4){3}}
               \put(24,14.5){\makebox(0,0){$_{t_8}$}}
             }
             \put(4,3){\line(1,0){1}}
             \put(4,2.6){\line(0,1){0.8}}
             \put(5,2.6){\line(0,1){0.8}}
             \put(5.6,0){\makebox(0,0){\scriptsize $\Delta_k$}}
             \put(16,3){\line(1,0){4}}
             \put(16,2){\line(0,1){1.8}}
             \put(20,2){\line(0,1){1.8}}
             \put(18,0){\makebox(0,0){\scriptsize $\delta_k$}}
           \end{picture}
         \end{minipage}
        \end{center}
       \end{small}
       \vspace*{.5cm}
    \caption{Example of successive trials in MADS.
    The grid in background is the mesh of coarseness $\Delta_k$ and the square around $\bphi_k$ is the poll frame of radius $\delta_k$.
    After each failure, the mesh and poll sizes ($\Delta_k$ and $\delta_k$) are reduced to
    allow more flexibility for new poll candidates.
    Figure adapted from~\cite{Le09b}.}
      \label{fig:mads_fails}
\end{figure}
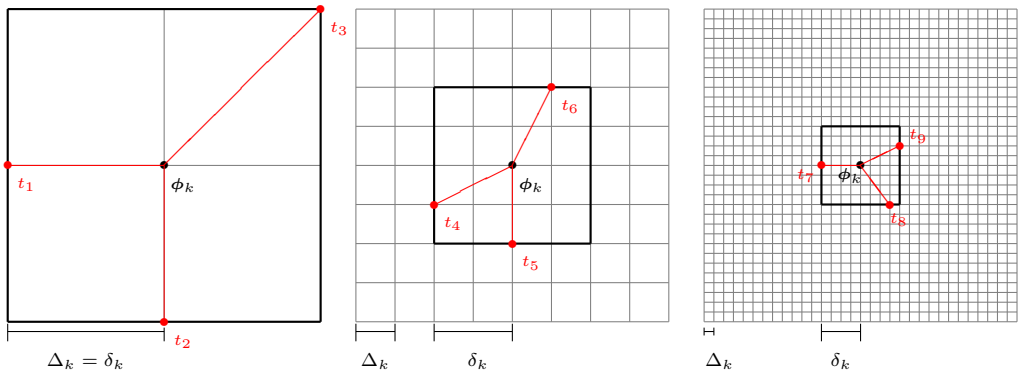

MADS starts the optimization process with an initial configuration $\bphi_0$, an initial mesh size $\Delta_0 \in \mathbb{R}$ and defines a mesh 
$M_k =  \left\{\bphi + \Delta_k I_n \z : \z \in \mathbb{Z}^n \right\}$, at each iteration $k$.
Each iteration is made of two steps: the {\em search} and the {\em poll}.  The {\em search} is flexible and may include various search space strategies as long as it produces a finite number of trial mesh points at each iteration. This phase usually implements a global search such as  \emph{Latin hypercube} sampling or surrogates which allow to explore further regions. 
The {\em poll} step is rigidly defined so that
global convergence to a local optimum is guaranteed under mild conditions.
The MADS algorithm generates a set of poll directions that grow asymptotically dense
around the incumbent $\bphi_k$ to produce a pool of local candidates around $\bphi_k$ at a radius of $\delta_k$, called the poll size.
These candidates are evaluated  {\em opportunistically}, so the poll stops as soon as a better solution is found and discards the remaining poll points.
An improvement on $\bphi_k$ is recorded if iteration $k$ is a success, in which case the mesh and poll sizes increase for the next iteration. Otherwise, the mesh and poll size are reduced while maintaining $\Delta_k \leq \delta_k$ as illustrated in Figure~\ref{fig:mads_fails}.




\section{Architecture Search}\label{sec:pb}

Following~\cite{pmlr-v97-tan19a} we define the architecture search over three parameters $\bphi=(\phi_1,\phi_2,\phi_3)$: the network depth $\phi_1$, the network width $\phi_2$ and the resolution $\phi_3$. The depth is searched by adding or removing ResNet blocks. The width is searched by modifying the output channels. The resolution is searched by changing the number of input channels.  

We start exploring the effectiveness of the {\sf NOMAD} solver by considering the unconstrained Problem~\ref{eq:nas_unconstrained} to maximize the validation accuracy on CIFAR-10 over a restrictive search space $\bphi\in\Phi$ that contains compressed architectures of ResNet-18. {\sf NOMAD} is compared against Microsoft Neural Network Intelligence {\sf NNI}, a standard mature hyperparameter tuning toolkit for neural networks. Practitioners use {\sf NNI} for hyperparameter tuning and also for architecture search.
\begin{figure}
    \centering
        \includegraphics[width=0.48\textwidth]{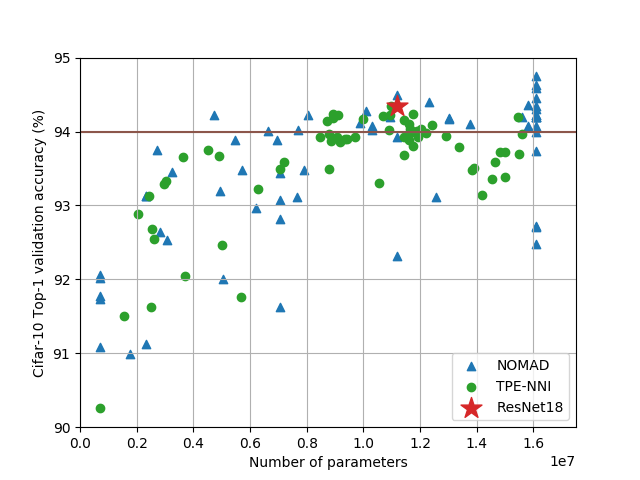}
        \includegraphics[width=0.48\textwidth]{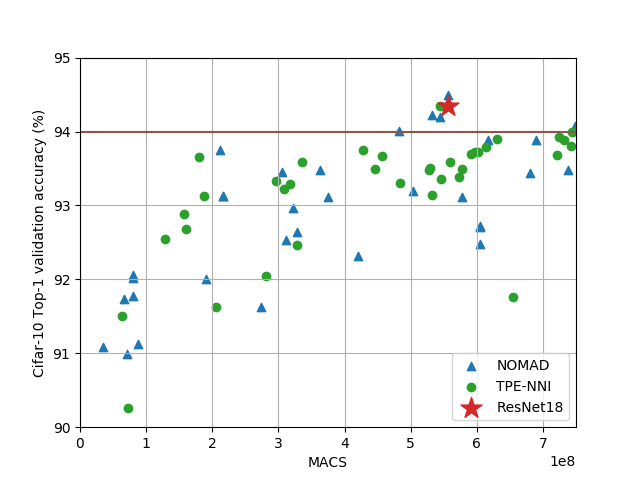}
    \caption{Searching for an architecture around ResNet-18 with maximum accuracy on CIFAR-10; top-1 accuracy on validation data versus number of network parameters $\times 10^7$ (left panel) and top-1 accuracy versus number of multiply-accumulate operations MACS (right panel). Networks with one third parameters and half MACS are found using {\sf NOMAD}. }
    \label{fig:cifar10-resnet18}
\end{figure}


 
The result of these experiments is shown in Figure~\ref{fig:cifar10-resnet18}. Starting from the ResNet-18 network that scores $94.5\%$, has $11\times 10^6$ parameters and $5.5\times 10^7$ MACS; both {\sf NOMAD} and Microsoft {\sf NNI} using its Bayesian optimization module {\sf TPE}~\citep{bergstra2011algorithms} are launched on CIFAR-10 with a budget of $100$ trials each. 

{\sf NOMAD} manages to find over $29$ configurations with a top-1 accuracy over $94\%$, which represents less than $0.35\%$ accuracy drop compared to the baseline. From these top performers, one solution has $4.8 \times 10^7$ MACS which represents a compression of $12.7\%$ and another solution has $4.7\times 10^6$ parameters that is equivalent to a $\times 2.34$ compression compared to the baseline. 

{\sf TPE-NNI} finds $19$ architectures with a top-1 accuracy over $94\%$. From these top performers, the one with lowest number of MACS has $5.4\times10^7$ MACS and the lowest number of parameters is recorded at $8.7\times10^6$ which represents a compression of $23\%$. 





\begin{figure}
    \centering
        \includegraphics[width=0.48\textwidth]{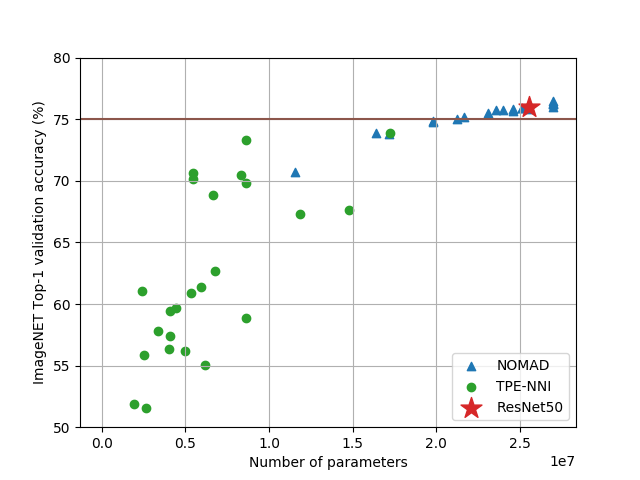}
        \includegraphics[width=0.48\textwidth]{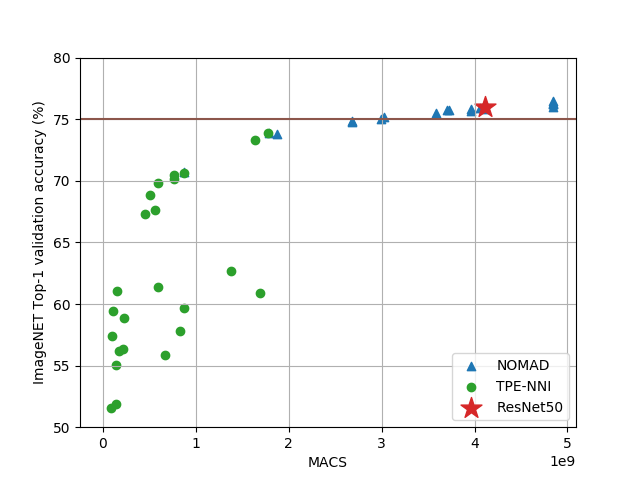}
    \caption{Searching for an architecture around ResNet-50 with maximum accuracy on ImageNet; top-1 accuracy on validation data versus number of network parameters $\times 10^7$ (left panel) and top-1 accuracy versus number of multiply-accumulate operations MACS (right panel).}
    \label{fig:imagenet-resnet50}
\end{figure}

A second test is performed where the neural architecture search is expressed with Formulation~\ref{eq:nas_constrained},
in which $\bphi_0$ is the ResNet-50 architecture and $\x,\y$ are the ImageNet training data. This experiment starts from the architecture of ResNet-50 which score a top 1 accuracy of $75.95\%$, has $\sim 25\times 10^6 $ parameters with $\sim 4\times 10^9$ MACS. This time, {\sf NOMAD} and
{\sf TPE-NNI} are allowed a budget of $25$ trials and the reported results are summarized in Figure~\ref{fig:imagenet-resnet50}. 

{\sf TPE-NNI} samples architectures with a high compression factor but at the cost of a significant drop in accuracy as the highest scoring candidate is still $2\%$ less accurate than the given baseline. This particular network has $\sim 17\times 10^6$ parameters and $\sim 1.8 \times 10^9$ MACS. The candidates found by the {\sf NOMAD} software have smaller compression ratios but maintains a good quality precision. Out of the $25$ sampled points, $15$ score higher than $75\%$ of which $2$ have at least $25\%$ less MACS than ResNet-50, {\sld
which is a clear improvement compared to {\sf NNI} with Bayesian optimization routine {\sf TPE}. }
Bayesian optimization has been shown successful in attacking blackbox problems effectively and ranked top at  blackbox optimization challenge recently~\citep{cowen2020hebo}. 

\section{Conclusion}

Traditionally deep learning community focuses on increasing accuracy on a benchmark data. Rapid use of deep models on edge devices calls for 
increasing accuracy while keeping resources to the minimal. Despite recent efforts in automating neural architectures, the vast majority of practical cases calls for manual design. Each device has its own constraints and recently developed automation tools ignore the constrained formulation of the problem. We showed the effectiveness of a derivative-free solver in two scenarios, when the search space embeds the constraint and the optimization is treated as an unconstrained problem, which is the common formulation.
In a second scenario, we minimized the computational resources while putting a hard constraint on model performance. 
This formulation is novel and requires solvers that take the constraint into account.
Our numerical results show that the derivative-free solver {\sf NOMAD} beats the popular Bayesian optimization {\sf TPE-NNI} framework.

\subsubsection*{Acknowledgments}
This research is supported by the NSERC Alliance grant 544900-19 in collaboration with Huawei-Canada.



\bibliography{main}
\bibliographystyle{iclr2021_conference}


\end{document}